\documentclass[conference]{IEEEtran}
\IEEEoverridecommandlockouts
\usepackage{cite}
\usepackage{amsmath,amssymb,amsfonts}
\usepackage{graphicx}
\usepackage{textcomp}
\usepackage{xcolor}
\usepackage{tablefootnote}
\usepackage{multirow}
\usepackage{amsmath}
\usepackage[ruled,vlined]{algorithm2e}

\def\BibTeX{{\rm B\kern-.05em{\sc i\kern-.025em b}\kern-.08em
    T\kern-.1667em\lower.7ex\hbox{E}\kern-.125emX}}
\begin{document}

\newcommand*{\arun}{\textcolor{blue}}
\newcommand*{\sidd}{\textcolor{red}}

\title{DeL-haTE: A Deep Learning Tunable Ensemble for Hate Speech Detection}

\author{\IEEEauthorblockN{Joshua Melton\IEEEauthorrefmark{1}, Arunkumar Bagavathi\IEEEauthorrefmark{2}, Siddharth Krishnan\IEEEauthorrefmark{1}}
\IEEEauthorblockA{\textit{Department of Computer Science} \\
\textit{University of North Carolina at Charlotte\IEEEauthorrefmark{1}, Oklahoma State University\IEEEauthorrefmark{2}}\\
jmelto30@uncc.edu, abagava@okstate.edu, skrishnan@uncc.edu}
}

\maketitle

\begin{abstract}
Online hate speech on social media has become a fast-growing problem in recent times. Nefarious groups have developed large content delivery networks across several mainstream (Twitter and Facebook) and fringe outlets (Gab, 4chan, 8chan, etc.) to deliver cascades of hate messages directed both at individuals and communities. Thus addressing these issues has become a top priority for large-scale social media outlets. Three key challenges in automated detection and classification of hateful content are the lack of clearly labeled data, evolving vocabulary and lexicon - hashtags, emojis, etc - and the lack of baseline models for fringe outlets such as Gab. In this work, we propose a novel framework with three major contributions. (a) We engineer an ensemble of deep learning models that combines the strengths of state-of-the-art approaches, (b) we incorporate a tuning factor into this framework that leverages transfer learning to conduct automated hate speech classification on unlabeled datasets, like Gab, and (c) we develop a weak supervised learning methodology that allows our framework to train on unlabeled data. Our ensemble models achieve an 83\% hate recall on the HON dataset, surpassing the performance of the state of the art deep models. We demonstrate that weak supervised training in combination with classifier tuning significantly increases model performance on unlabeled data from Gab, achieving a hate recall of 67\%.
\end{abstract}


\section{Introduction}

The proliferation of online hate speech has become prevalent in recent times. Numerous social media outlets and the computational social science community are looking at various automated techniques to detect and classify hate speech. However, most models, nascent in nature, have significant limitations due to the complexity of the problem. Primarily, the lack of a reliable baseline coupled with an evolving vocabulary of hateful content makes this a particularly challenging issue. For instance, many studies have classified this problem as a binary classification task~\cite{zhang2018hate, watanabe2018hate}, but this fails to address the subtleties of hate speech, such as direct (use of vulgar language against an individual or community) vs. indirect (the content can be lexicographically clean but implies negative content) hate speech. These binary classification models also fail to identify different types of hate speech like racism, sexism, antisemitism, etc. or their varying degrees. Another key obstacle that plagues these binary models is their inability to distinguish between general offensive language and hate speech~\cite{davidson2017hon}. A third issue that arises in designing automated approaches is class imbalance---hate speech is usually a small percentage of the overall data---and the need to adequately upsample hate observations without model overfitting.

In our work, inspired by the recent successes in developing multi-class hate speech models that separate hate speech from offensive content~\cite{davidson2017hon, rizos2019augment}, we propose \emph{DeL-haTE}, an ensemble of tunable deep learning models that leverages CNN and GRU layers. The CNN layer extracts higher-order features from the word embedding matrix that then inform the GRU layer, which extracts informative features from the sequence of words. These features are utilized for automatic detection of hate speech on social media. Our novelty lies in using a tuning procedure to adapt the model to individual dataset characteristics. 

%
%


Our major contributions can be summarized by answering the following questions.

\begin{enumerate}
	\item{\textbf{How do you leverage existing state of the art classification models for automatic hate speech detection?} We utilize existing deep model topologies to develop an ensemble classifier model for hate speech detection. An ensemble approach effectively tackles issues of class imbalance and model variability that are significant problems for automatic detection of hate speech.}
	\item{\textbf{How can you engineer a generalized framework for hate speech detection on unlabeled data?}}
	\begin{enumerate}
		\item{\textbf{How can you extend trained classification models to evolving, unlabeled data?} We extend pretrained models by applying transfer learning to tune the classifiers to new target datasets. The tunability of our framework allows the model to adapt to new and ever-changing data.}
		\item{\textbf{How can you develop an unsupervised framework for unlabeled data?} We develop a weak supervision methodology that allows our framework to train and tune entirely on unlabeled data, further extending the applicability of our model to new data.}
	\end{enumerate}
	
\end{enumerate}

\textbf{Summary of Results:} Our best ensemble on the HON dataset achieves a 65\% F1 Macro and an 83\% hate recall, surpassing the performance on the HON dataset of current state of the art models by 33\%. We show that the ensemble models outperform individual models by an average of 5\% hate recall and 8\% F1 macro across all datasets. When applied to unlabeled Gab data, tuning improved the pretrained models by an average of 12\%, with the best tuned ensemble models achieving 57\% hate recall. Our model trained using weak supervision achieved a 67\% hate recall on posts from Gab.

%
%

\section{Related Work}

Developing a consistent definition of \emph{hate speech} is difficult due to its controversial and subjective nature~\cite{zannettou2018gab,macavaney2019hate}. Social media sites define hate speech in legal terms as a “\emph{direct attack}” or “\emph{promoting violence}” against various characteristics of people, including race, ethnicity, nationality, religion, gender, and others. Many previous analyses have approached the study of hate speech analysis through the lenses of these characteristics. Examples include automatic hate speech detection modeled as - a binary classification problem~\cite{zhang2018hate, watanabe2018hate}, an attention-based multi-task learning model to identify toxic comments~\cite{vaidya2020empirical}, a quantification of conflicting opinion among communities~\cite{garimella2018political}, and a racism/sexism classifier with embeddings learned from multiple deep learning architectures~\cite{badjatiya2017deep}. A binary classification based approach, although simple, ignores the many subtleties of hate speech, such as indirect vs. direct hate speech and different forms of hate speech, such as racism, sexism, or antisemitism. Furthermore, the high prevalence of offensive language on social media presents an additional challenge for automatic hate speech detection online~\cite{davidson2017hon}.

%

Several multi-class classification models~\cite{davidson2017hon, zhang2018hate, rizos2019augment, arango2019hate} have been introduced recently to better distinguish hate speech from offensive content on social media and to improve the automatic identification of various types of hate speech. In a similar vein, hierarchical annotation systems have been proposed that further distinguish the type and target of offensive posts, providing additional granularity for automatic detection models~\cite{zampieri2019olid}. One of the major issues with automatic hate speech detection research is the limited amount of manually labeled data. Weak supervised training allows for the use of large unannotated datasets by programmatically generating ``weak" labels using heuristic approaches\cite{erhan2010weak, dehghani2017weak}. Weak supervised learning has been applied for problems like cyberbullying detection to give better performance than traditional approaches~\cite{raisi2018weak}. 



Despite the strong performance of recent automatic hate speech detection models, with high reported recall and F1 scores (above 90\%), efforts to replicate reported findings and to generalize models to other similar datasets have often failed~\cite{arango2019hate, grondahl2018evade}. While some of these failings are due to methodological shortcomings such as overfitting, they are also related to the inherent subjectivity of hate speech, the noisiness of short-text social media posts, and the biases present in datasets~\cite{arango2019hate}. With binary classification, it is also difficult to report the extent to which hate speech detection models are conflating hate speech with general offensive speech online~\cite{davidson2017hon}. In addition to these challenges, hate speech constitutes a small portion of the overall content on social media~\cite{zannettou2018gab, lima2018gab}, leading to the presence of severe class imbalance in hate speech datasets. Example include the datasets we use in this work as given in Table~\ref{tab:data_dist}. Despite recent efforts to identify and address these challenges~\cite{zhang2018hate, rizos2019augment, arango2019hate}, there remains room for improvement in developing robust and generalized frameworks for automatic detection of hate speech on social media.

In this work, we develop a number of multi-class classifiers comparing four sets of pretrained word embeddings and three different deep model architectures. We leverage an ensemble approach to address issues of class imbalance and the limited number of per-class observations during model training. In order to assess the generalizability of our model frameworks, we apply transfer learning to tune classifiers trained on existing labeled Twitter datasets and test using a small sample of manually labeled posts from Gab.ai. 



\section{Datasets}


We use two Twitter datasets for our experiments, which are referred to as HON~\cite{davidson2017hon} and OLID~\cite{zampieri2019olid} throughout this paper. We also use unlabeled posts from the social media forum called \emph{gab.com} or \emph{gab.ai}. \emph{gab.com} started gaining traction immediately after the 2016 U.S. Presidential election due to its support for \emph{free speech} in online media. Much previous research has showcased evidence of the spread of antisemitic and racist ideologies in this forum~\cite{lima2018gab, zannettou2018gab}. We utilize this data to examine the generalizability of our developed framework, and the dataset is referred to as Gab throughout this paper. For the HON dataset, Davidson et al.~\cite{davidson2017hon} classified tweets into three categories: hate speech, offensive language, and neither. In order to standardize our experimental setup, we apply this data representation to all the above mentioned datasets in our analysis. 


The HON dataset\footnote{https://github.com/t-davidson/hate-speech-and-offensive-language} consists of approximately 25,000 labeled tweets. These tweets were originally sampled using the hate speech lexicon from \emph{Hatebase.org}\footnote{https://hatebase.org}, which identifies common words and phrases that are marked as hate speech by online users. This data corpus is a random sample the complete sets of tweets from selected users (8.54 million). This sample was then manually labeled using \emph{CrowdFlower} to produce the final labeled dataset~\cite{davidson2017hon}.

The OLID dataset\footnote{https://scholar.harvard.edu/malmasi/olid} consists of about 13,000 labeled tweets. This data corpus was collected from Twitter using a set of keywords and constructions that are often included in offensive tweets, with main emphasis on political keywords that are more likely to result in offensive content. These tweets are then manually labeled using \emph{Figure Eight} as \emph{offensive} and \emph{not offensive}. Each tweet in this dataset is annotated with a three level hierarchical scheme: denoting offensive language, the category of offensive language, and the target of offensive language~\cite{zampieri2019olid}. For our experiments, we adapt the OLID hierarchical labels into the (H)ate, (O)ffensive, (N)either three-class labels by labeling offensive posts that are targeted at a group as \textbf{H}ate, remaining offensive posts as \textbf{O}ffensive, and not offensive posts as \textbf{N}either.



The Gab dataset consists of approximately 1,500 posts that were randomly sampled from a Gab posts database containing over 35 million posts. These posts were manually annotated by two researchers utilizing the procedure described by \cite{davidson2017hon}. For weak supervised training, a randomly sampled set of nearly 100,000 unlabeled posts was utilized.

\begin{table}
	\renewcommand*{\arraystretch}{1.2}
	\centering
	
	\caption{Distribution of Hateful (H), Offensive (O), and Neither (N) data samples in multiple datasets. Hateful posts are only 5\%-11\% of the total corpus for machine learning training}
	\begin{tabular}{ccccc}
		\hline
		\textbf{Dataset} & \multicolumn{3}{c}{\textbf{Class (in \%)}} & \textbf{\# Samples} \\
		& H & O & N & \\
		\hline
		HON & 5.74 & 77.41 & 16.85 &  22,305 \\
		OLID & 8.24 & 25.09 & 66.67 & 11,916 \\
		Combined & 6.61 & 59.19 & 34.20 & 34,221 \\
		\hline
		Gab - Test & 11.19 & 22.05 & 66.76 & 1,465 \\
		Gab - Train & -- & -- & -- & 90,899 \\
		\hline
	\end{tabular}
	\label{tab:data_dist}
\end{table}

\begin{figure*}
	\centering
	\includegraphics[scale=0.5]{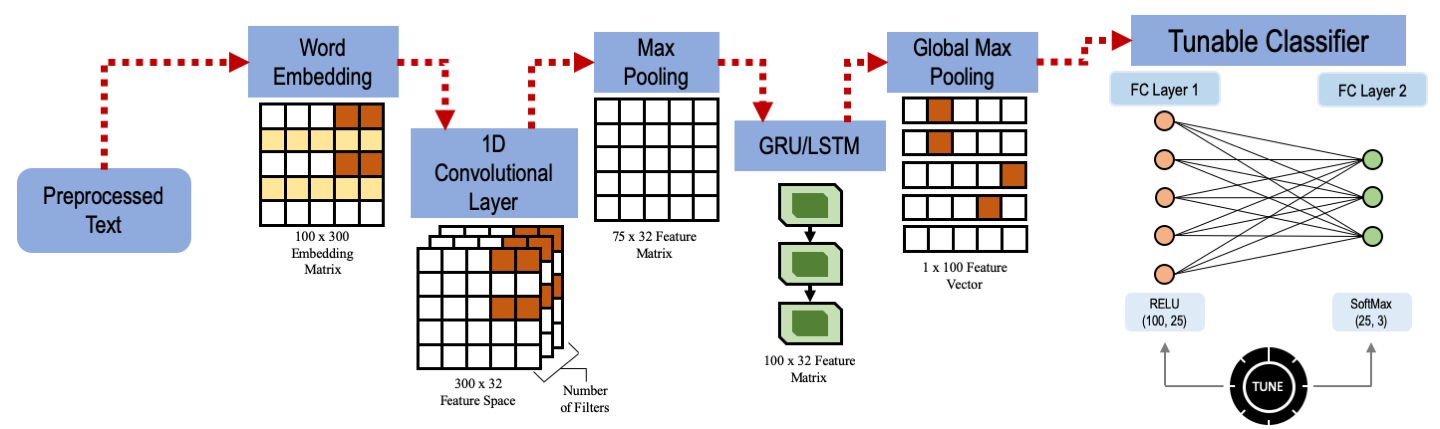}
	\caption{Overview of the entire pipeline process for a single post. The model topology depicted illustrates the CNN/GRU feature component and the tunable classifier. These models are combined into an ensemble model for class prediction.}
	\label{fig:model_pipeline}
\end{figure*}

\section{Methodology}
The code for this paper is available on Github\footnote{https://github.com/NASCL/DeL-haTE}. We conduct a comparative analysis of three deep model architecture variants in order to determine the optimal deep model architecture and word embedding representation. We compare the following five word embedding methods: Word2Vec vectors trained on Google News corpus~\cite{mikolov2013w2v}, GloVe vectors trained on CommonCrawl (GLoVe-CC) and Twitter (GLoVe-Twitter) corpora \cite{pennington2014glove}, and FastText vectors trained on CommonCrawl (FastText-CC) and Wikipedia (FastText-Wiki) corpora \cite{bojanowski2016fasttext}. The pre-trained distributional embedding vectors are all implemented using PyTorch's torchtext.Vocab library. To examine the issues of class imbalance, limited labeled training data, and model generalizability, we develop an ensemble model approach and compare transfer learning and weak supervised training using data from Gab. For our experiments, we report Macro F1, which weights each class equally regardless of size, and hate class recall. All reported results are averaged across 5 trials with an ensemble size of 5 models.

\subsection{Text Processing}
We follow a standardized procedure to preprocess all posts in all datasets that are used in our experiments. In the preprocessing, we remove all extra text elements like URLs and emojis. User mentions are normalized to "MENTIONHERE", and hashtags are normalized to "HASHTAGHERE \textless hashtag\_text\textgreater". We then apply tokenization and stemming to the normalized texts using NLTK's word tokenizer and Porter Stemmer, respectively. We retrieve word embeddings for the stemmed tokens with pre-trained models: \emph{Word2vec}, \emph{GloVe}, and \emph{FastText}. We standardize the size of word embedding with zero left padding to convert the word embeddings matrix dimension to $100 \times 300$ for \emph{word2vec}, \emph{GloVe-CC}, and both \emph{FastText} models and $100 \times 200$ for the \emph{GloVe-Twitter} model.

\subsection{Deep Model Implementation}

Our deep learning framework is motivated from the deep model topologies proposed in~\cite{zhang2018hate} and~\cite{rizos2019augment}. A schematic of our model is given in Figure~\ref{fig:model_pipeline}. A 1-dimensional CNN layer takes the $100 \times 300$ word embedding matrix as input. We utilize a CNN layer with 32 filters and a filter width of 17 with padding of 8 to match input and output dimensions. We then apply max pooling with an undersampling rate of 4 to generate a $75 \times 32$ feature space. With this vector space, we compare between two sub-models using combinations of convolutional (CNN), recurrent (RNN), and fully connected (FC) layers:

\begin{itemize}
	\item \textbf{CNN-RNN-FC}: For the CNN-RNN-FC model variant, the $75 \times 32$ feature space is passed to RNN layer with 100 hidden units. We use an LSTM or GRU as the RNN layer in our experiments. The $100 \times 32$ output is flattened to a 3,200-dimension vector, and global max pooling is applied to generate a 100-dimension feature vector. The 100-dimension feature vector is passed to an FC layer that utilizes ReLU activation and outputs 25 hidden units. For the hidden FC layer, we use dropout with a probability of 0.2 during training. A final output FC layer utilizes Softmax activation and outputs class probabilities for \textbf{H}ate, \textbf{O}ffensive, and \textbf{N}either.
	\item \textbf{CNN-FC}: For the CNN-FC model variant, the $75 \times 32$ output feature space from the CNN layer is flattened to a 2,400-dimension feature vector and passed to the two FC layers.
\end{itemize}  

\subsection{Ensemble Training}
We follow ensemble training with five independent versions of our developed CNN-RNN-FC model variants discussed in the previous section. We use a standard data split of 80-10-10 train-valid-test for the ensemble training, tuning, and testing. For each ensemble training epoch, we pass equal class distributions of hate, offensive, and neither. Due to high frequency of \emph{Offensive} and \emph{Neither} classes, we randomly sample \emph{Offensive} and \emph{Neither} class data with replacement. We run 20 epochs of training, utilizing early stopping to save the model weights and biases at the epoch with the minimum validation loss. The prediction from the ensemble classifier is the majority decision from five independent classifiers. 



\subsection{Transfer Learning}
To test the generalizability of our framework's hate speech prediction, we experiment with transfer learning by training the classifiers on Twitter-based HON, OLID, and Combined datasets and evaluating on Gab posts. As reported in Section \ref{tl}, we find that the models tested on completely unseen data samples give low performance. Thus we tune classifiers trained on our labeled experiment datasets with a small, manually labeled set of posts from Gab. A set of 150 posts (50 posts for each class Hate, Offensive, and Neither) is selected as a balanced training set for transfer learning. For the tuning, we freeze the weight and bias parameters of feature extraction CNN-RNN layers, while the classifier component with the two FC layers, is allowed to train on the Gab data. The exceptionally small size of the training data makes the models highly sensitive to overfitting on the Gab observations. For tuning the models' classifier components, the training process runs for 10 epochs at a reduced learning rate of $5\mathrm{e}{-4}$.


\subsection{Weak Supervised Training}

\begin{figure}
	\centering
	\includegraphics[scale=0.5]{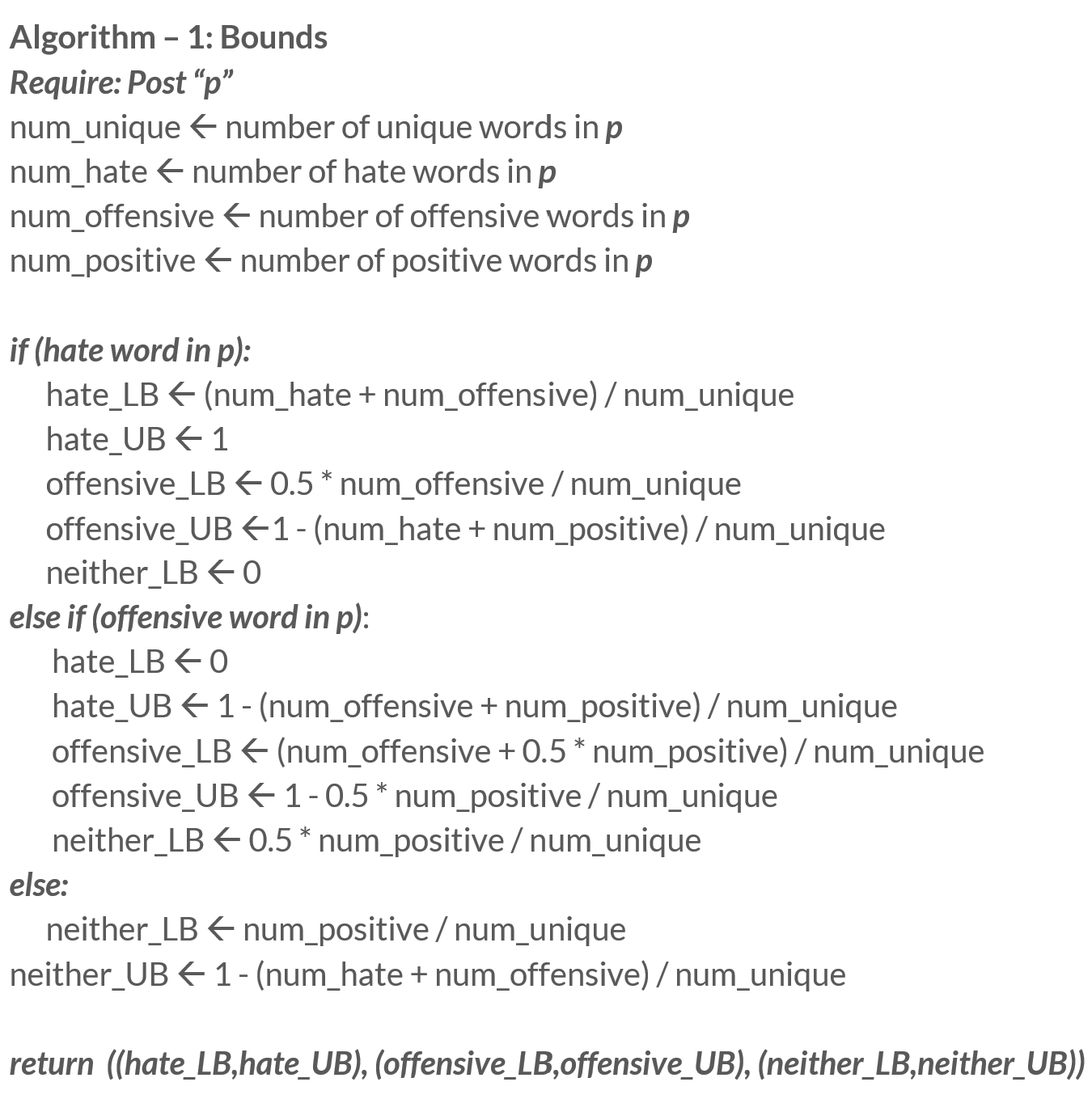}
	\caption{Weak supervised bounds algorithm that calculates a lower and upper bound for each class.}
	\label{fig:algo}
\end{figure}

Utilizing weak supervision during model training is a possible alternative to reliance on manually annotated hate speech datasets. The major benefit of weak supervision is the greatly expanded pool of training observations by leveraging abundant unlabeled social media data. For our classifier models, the framework utilizes a weakly supervised form of cross-entropy as the loss function $f$. This loss function is the multi-class extension of the weak supervised loss function utilized by~\cite{raisi2018weak} for binary classification of cyberbullying. Our weak supervised loss relies on lexicons of words indicative and counter-indicative of hate speech and offensive language. For a post containing $n$ unique words, let $n_h$, $n_o$, and $n_p$ denote the number of hate words, offensive words, and positive words, respectively. For each post, we calculate the bounds of each class using the algorithm described in Figure~\ref{fig:algo}.


If this bound is violated, the loss function penalizes the model using the weak supervised loss function:

\begin{equation*}
\begin{aligned}
	f(y_m) = \sum_{c \in C} (-log(min\{1, 1 + y_{m,c} - lb_{m,c}\}) \\
		- log\{min(1, 1 + ub_{m,c} - y_{m,c}\})) * w_c
\end{aligned}
\end{equation*}

With the lack of \textit{a priori} class labels, this form of weak supervised training is susceptible to the class imbalance problem due to the small proportion of hate speech in the overall content on social media. To mitigate this issue, a per-class weight $w_c$ is applied to the weak supervised loss contribution for each class. One benefit of our weak supervised heuristic is the tunable per-class weight applied during the loss calculation which can be adjusted for varying class balances in datasets.

\section{Results}

In this section, we present the findings from our experiments on ensemble training, transfer learning, and weak supervision for automatic hate speech detection. With our experiments we show that (i) our ensemble approach outperforms the state of the art deep models on the HON dataset, (ii) our transfer learning with classifier tuning improves performance of pre-trained models on a novel dataset, and (iii) our weak supervision methodology can train our framework on entirely unlabeled data.

\subsection{Surpassing Performance on the HON Dataset}

We first focus on extending and improving the CNN-RNN-FC model architecture and surpassing the performance of the current best models on the HON dataset (Table~\ref{tab:hon_literature}). The CNN-RNN-FC toplogy with a single 1-dimensional convolutional layer and an LSTM/GRU layer has been shown to be an effective model for hate speech classification~\cite{rizos2019augment, zhang2018hate}. We adapt this model to develop an ensemble classifier utilizing both the HON and OLID datasets. Table~\ref{tab:ensembles} summarizes our findings on the Twitter datasets. The model variant trained on HON consisting of word embeddings from a pre-trained FastText model trained on a Wikipedia corpus in combination with a CNN+GRU architecture achieves F1 Macro of 65\% and a hate recall of 83\%, outperforming the state of the art deep models on the HON dataset by 33\% on hate recall. 

\begin{table}
	\renewcommand*{\arraystretch}{1.3}
	\centering

	\caption{Baseline classification models applied to the HON dataset. Reported results are from the original corresponding studies.}
	\begin{tabular}{ccc}
		\multicolumn{3}{c}{\textbf{HON - Baseline}} \\
		\textbf{Topology} & \textbf{F1-Micro} & \textbf{Hate Recall} \\
		\hline
		LogRegBase~\cite{davidson2017hon} & 0.91 & 0.61 \\
		DeepBase~\cite{zhang2018hate} & 0.94* & n/a* \\
		\hline
		& \textbf{F1-Macro} & \textbf{Hate Recall} \\
		\textit{GloVe-CC}+CNN+LSTM+BestAug~\cite{rizos2019augment} & 0.741 & 0.496\\
		\hline
		\multicolumn{3}{c}{* Binary classification task - hate recall was not reported.}	
	\end{tabular}
	\label{tab:hon_literature}
\end{table}

\begin{table*}
	\renewcommand*{\arraystretch}{1.3}
	\centering

	\caption{Comparison of the network topologies on the \emph{HON}, \emph{OLID}, and \emph{Combined} datasets. Each block corresponds to a different word embedding and reports the results for the three deep model variants.}
	\begin{tabular}{c|cc|cc|cc}
		& \multicolumn{2}{c|}{\textbf{HON Ensemble}} & \multicolumn{2}{c|}{\textbf{OLID Ensemble}} & \multicolumn{2}{c}{\textbf{Combined Ensemble}} \\
		\textbf{Topology} & \textbf{F1-Macro} & \textbf{Hate Recall} & \textbf{F1-Macro} & \textbf{Hate Recall} & \textbf{F1-Macro} & \textbf{Hate Recall}\\
		\hline
		\textit{Word2Vec}+CNN+FC & 0.63 & 0.47  & 0.48 & 0.16 & 0.63 & 0.33 \\
		\textit{Word2Vec}+CNN+GRU & 0.57 & 0.59 & 0.43 & 0.14 & 0.61 & 0.45 \\
		\textit{Word2Vec}+CNN+LSTM & 0.60 & 0.50  & 0.44 & 0.16 & 0.61 & 0.44  \\
		\hline
		\textit{GloVe-Twitter}+CNN+FC & 0.66 & 0.53 & \textbf{0.35} & \textbf{0.44} & 0.64 & 0.46 \\
		\textit{GloVe-Twitter}+CNN+GRU & 0.65 & 0.78 & 0.47 & 0.33 & 0.66 & 0.65 \\
		\textit{GloVe-Twitter}+CNN+LSTM & 0.64 & 0.53 & 0.42 & 0.19 & 0.63 & 0.46 \\
		\hline
		\textit{GloVe-CC}+CNN+FC & 0.68 & 0.44 & 0.48 & 0.38 & 0.65 & 0.55 \\
		\textit{GloVe-CC}+CNN+GRU & 0.65 & 0.48 & 0.54 & 0.41 & \textbf{0.65} & \textbf{0.66} \\
		\textit{GloVe-CC}+CNN+LSTM & 0.67 & 0.48 & 0.48 & 0.23 & 0.66 & 0.46 \\
		\hline
		\textit{FastText-CC}+CNN+FC & 0.70 & 0.53 & 0.47 & 0.31 & 0.67 & 0.50 \\
		\textit{FastText-CC}+CNN+GRU & 0.68 & 0.74 & 0.51 & 0.36  & 0.70 & 0.62 \\
		\textit{FastText-CC}+CNN+LSTM & 0.68 & 0.55 & 0.42 & 0.20 & 0.65 & 0.51 \\
		\hline
		\textit{FastText-Wiki}+CNN+FC & 0.68 & 0.54 & 0.47 & 0.36 & 0.64 & 0.54 \\
		\textbf{\textit{FastText-Wiki}+CNN+GRU} & \textbf{0.65} & \textbf{0.83} & 0.56 & 0.36 & 0.67 & 0.65 \\
		\textit{FastText-Wiki}+CNN+LSTM & 0.65 & 0.52 & 0.43 & 0.24 & 0.65 & 0.46 \\
		\hline
	\end{tabular}
	\label{tab:ensembles}
\end{table*}


Due to the smaller size of OLID dataset---with only half the size of the HON dataset---and with imperfect mapping of the original hierarchical labels to the hate, offensive, and neither labels, Table~\ref{tab:ensembles} illustrates that performance on the OLID dataset is lower than the HON dataset. Our primary motivation in utilizing the OLID dataset is to augment the available training data and to improve the model's generalizability to new datasets. 

Integrating the HON and OLID datasets into a single training set increases both the size (a nearly 50\% increase from the HON dataset alone) and the diversity of training examples for the model to learn from during training (Table~\ref{tab:data_dist}). The increased training pool also improves the downsampling procedure utilized to mitigate the class imbalance of hate speech datasets. The size of the train set passed to the model at each epoch is limited by the size of the minority class. The Combined dataset contains an additional 1,000 observations in the hate class. Table~\ref{tab:ensembles} presents our findings for the ensemble models trained on the Combined HON+OLID dataset. We find that the F1-Macro of these models is able to equal the performance of the models trained on HON alone.

\subsection{Benefits of the Ensemble Approach}

A major factor contributing to the variability in hate speech detection model performance is the severe class imbalance present in hate speech datasets~\cite{arango2019hate,grondahl2018evade}. Similar to~\cite{rizos2019augment}, we likewise found that using downsampling during training is critical in preventing the model from almost exclusively predicting the majority class. One drawback of downsampling during training is overfitting on the smaller distribution of hate class examples during training. In addition, random sampling of observations in the remaining classes can lead to variability in the resultant models. We therefore employ a simple ensemble approach~\cite{zimmerman2018ensemble} to counteract the variability in individual classifier models for hate speech detection.


Our experiments demonstrate that utilizing an ensemble approach for hate speech classification significantly improves model performance when compared against individual classifier models. In Table~\ref{tab:indiv_ensemble}, we summarize our findings regarding the benefits of an ensemble approach over individual models. The individual results are averaged over the 25 component models, and the ensemble results are averaged over five trials with an ensemble size of five. On average, the ensemble models outperform individual models by 5\% in hate recall and 8\% in F1 Macro.



\begin{table*}
	\renewcommand*{\arraystretch}{1.3}
	\centering

	\caption{Comparison of the best-performing ensemble models versus their individual component models. Ensemble models outperform their component models by an average of 8\% in F1 Macro and 5\% in Hate recall.}
	\begin{tabular}{cccccc}
		\multicolumn{6}{c}{\textbf{Individual vs. Ensemble Models}} \\
		& & \multicolumn{2}{c}{\textbf{Individual}} & \multicolumn{2}{c}{\textbf{Ensemble}} \\
		\textbf{Dataset} & \textbf{Topology} & \textbf{F1-Macro} & \textbf{Hate Recall} & \textbf{F1-Macro} & \textbf{Hate Recall} \\
		\hline
		\multirow{4}{*}{HON} & \textit{GloVe-Twitter}+CNN+GRU & 0.55 & 0.69 & 0.65 & 0.78 \\
			& \textit{GloVe-CC}+CNN+LSTM & 0.62 & 0.53 & 0.67 & 0.48 \\
			& \textit{FastText-CC}+CNN+GRU & 0.54 & 0.61 & 0.68 & 0.74 \\
			& \textbf{\textit{FastText-Wiki}+CNN+GRU} & \textbf{0.52} & \textbf{0.72} & \textbf{0.65} & \textbf{0.83} \\
		\hline
		\multirow{4}{*}{OLID} & \textbf{\textit{GloVe-Twitter}+CNN+FC} & 0.31 & 0.42 & 0.35 & 0.44 \\
			& \textit{GloVe-CC}+CNN+GRU & 0.44 & 0.41 & 0.54 & 0.41 \\
			& \textit{FastText-CC}+CNN+GRU & 0.42 & 0.34 & 0.51 & 0.36 \\
			& \textit{FastText-Wiki}+CNN+GRU & 0.46 & 0.35 & 0.56 & 0.36 \\
		\hline
		\multirow{4}{*}{Combined} & \textit{GloVe-Twitter}+CNN+GRU & 0.60 & 0.59 & 0.66 & 0.65 \\
			& \textbf{\textit{GloVe-CC}+CNN+GRU} & 0.58 & 0.63 & 0.65 & 0.66 \\
			& \textit{FastText-CC}+CNN+GRU & 0.62 & 0.58 & 0.70 & 0.62 \\
			& \textit{FastText-Wiki}+CNN+GRU & 0.60 & 0.59 & 0.67 & 0.65 \\
	\end{tabular}
	\label{tab:indiv_ensemble}
\end{table*}

In our experiments, we found that the GRU variant of the CNN-RNN topology generally outperformed the other model variants. For the HON dataset, GRU models are on average 17\% better on hate recall and are equal in average F1 Macro when compared with the other variants across all embedding choices. A similar trend is seen for the OLID and Combined datasets, where the GRU variants also outperformed the other variants on F1 Macro. Our experiments indicate that the GRU model variant may be better suited to extracting informative patterns from sequential features for hate speech classification tasks. Because of the improved performance, training times, and computational resource limitations, we elected to conduct all experiments with weak supervision using the CNN-GRU-FC model variant.

\begin{table*}
	\renewcommand*{\arraystretch}{1.3}
	\centering

	\caption{Comparison of pre-trained ensembles evaluated on the \emph{Gab} dataset. Results are reported before and after tuning is applied to the models. Tuned ensembles showed an average improvement of 12\% on Hate recall.}
	\begin{tabular}{cccccc}
		\multicolumn{6}{c}{\textbf{Pretrained Ensembles - Gab}} \\
		& & \multicolumn{2}{c}{\textbf{Pre-tuning}} & \multicolumn{2}{c}{\textbf{Post-tuning}} \\
		\textbf{Train Data} & \textbf{Topology} & \textbf{F1-Macro} & \textbf{Hate Recall} & \textbf{F1-Macro} & \textbf{Hate Recall} \\
		\hline
		\multirow{4}{*}{HON} & \textbf{\textit{GloVe-Twitter}+CNN+GRU} & 0.38 & 0.24 & 0.33 & 0.50 \\
			& \textit{GloVe-CC}+CNN+LSTM & 0.42 & 0.17 & 0.42 & 0.34 \\
			& \textit{FastText-CC}+CNN+GRU & 0.42 & 0.25 & 0.41 & 0.39 \\
			& \textit{FastText-Wiki}+CNN+GRU & 0.40 & 0.35 & 0.31 & 0.44 \\	
		\hline
		\multirow{4}{*}{OLID} & \textbf{\textit{GloVe-Twitter}+CNN+FC} & 0.39 & 0.43 & 0.31 & 0.43 \\
			& \textit{GloVe-CC}+CNN+GRU & 0.46 & 0.19 & 0.36 & 0.23 \\
			& \textit{FastText-CC}+CNN+GRU & 0.39 & 0.08 & 0.33 & 0.22 \\	
			& \textit{FastText-Wiki}+CNN+GRU & 0.44 & 0.11 & 0.39 & 0.20 \\
		\hline
		\multirow{4}{*}{Combined} & \textit{GloVe-Twitter}+CNN+GRU & 0.43 & 0.41 & 0.41 & 0.50 \\
			& \textit{GloVe-CC}+CNN+GRU & 0.45 & 0.40 & 0.41 & 0.50 \\
			& \textit{FastText-CC}+CNN+GRU & 0.45 & 0.27 & 0.46 & 0.44 \\
			& \textbf{\textit{FastText-Wiki}+CNN+GRU} & \textbf{0.46} & \textbf{0.44} & \textbf{0.40} & \textbf{0.57} \\
	\end{tabular}
	\label{tab:gab_tuning}
\end{table*}
		
\subsection{Extension to Gab Data: Transfer Learning} \label{tl}

The use of transfer learning for NLP tasks, and particularly for automatic hate speech detection, remains an open research question~\cite{semwal2018transfer, howard2018tuning}. Our experiments on the HON and OLID datasets demonstrate the success of our ensemble CNN-GRU-FC model architecture on curated, labeled datasets. In order to examine how well our model can be extended to novel, unlabeled data from a non-Twitter social media source, we conduct a series of experiments applying transfer learning tuning using the Gab dataset. We utilize the labeled HON and OLID data as source datasets to train classifier ensembles and then tune these pretrained models using data from Gab as the target dataset.


Table~\ref{tab:gab_tuning} summarizes our findings when applying models trained on the HON and OLID datasets to Gab data. The overall reduction in performance is in line with similar decreases in performance found by~\cite{arango2019hate} when applying hate speech classification models to novel datasets. The pre-trained models are evaluated on the manually labeled Gab test set prior to and after tuning, demonstrating an average 12\% improvement in hate recall across all models on the Gab test set after tuning. The best performing model is trained on the Combined HON+OLID dataset and after tuning, achieved a hate recall of 57\%. These results indicate that tuning of classifier models can lead to significantly improved model performance when applying automatic hate speech detection models to novel datasets.

As we hypothesized, the models trained on the Combined dataset generalize better to the novel Gab dataset. After tuning, ensembles trained on the Combined dataset outperform models trained on the HON dataset alone by an average of 10\% on hate recall and 5\% on F1 Macro. On average, the ensembles trained on the Combined dataset and then tuned to Gab data achieved a hate recall equivalent to that of the state of the art models on the HON dataset (Table ~\ref{tab:hon_literature}). Our experiments demonstrate the generalizability of our ensemble framework and indicate that tuning classifier models trained on well-annotated datasets improves the performance of models when extended to novel data.


\subsection{Extension to Gab Data: Weak Supervision}

Manually labeled datasets, such as HON and OLID, are a valuable asset for research on hate speech on social media, but these datasets constitute only a minute portion of the total content on social media. There remains a need for robust metholodogies that can successfully utilize large amounts of unannotated social media data. Besides transfer learning, weak supervision methods that create a set of ``weak" labels using heuristics allow for the use of unannotated data for training machine learning models~\cite{erhan2010weak, dehghani2017weak}. In our experiments, we develop a heuristic method for the Hate, Offensive, and Neither scheme that generates a per-class bound for each class. We test our procedure on the HON and OLID datasets, as well as on a large unannotated corpus of posts from Gab.

Table~\ref{tab:weak} summarizes our finding regarding weak supervised training using the HON and OLID datasets. The weak supervised models do not equal the performance of the models trained using the dataset labels. But, when tuned and tested on the novel Gab data, the weak supervised ensembles are able to equal or surpass the performance of the standard ensembles. Our experiments demonstrate a marked improvement in hate recall, 18\% on average, and in F1 Macro, 3\% on average. Table~\ref{tab:weak_gab} summarizes our findings for the weak supervised ensembles when evaluated on the Gab dataset. Our experiments demonstrate that while weakly supervised models fail to equal the performance of standardly trained models on labeled datasets, these ensemble models can equal and exceed their performance on novel data, such as the data from Gab.


\begin{table}
	\renewcommand*{\arraystretch}{1.3}
	\centering

	\caption{Ensemble classifier models trained on the HON and OLID datasets using weak supervision instead of the dataset labels. Results are reported for each model evaluated on its respective dataset.}
	\begin{tabular}{cccc}
		\multicolumn{4}{c}{\textbf{Weak Supervised Ensembles}} \\
		\textbf{Train Data} & \textbf{Topology} & \textbf{F1-Macro} & \textbf{Hate Recall}\\
		\hline
		\multirow{4}{*}{OLID} & \textit{GloVe-Twitter}+CNN+GRU & 0.29 & 0.24 \\
			& \textbf{\textit{GloVe-CC}+CNN+GRU} & 0.31 & 0.27 \\
			& \textit{FastText-CC}+CNN+GRU & 0.29 & 0.15 \\
			& \textit{FastText-Wiki}+CNN+GRU & 0.30 & 0.19 \\	
		\hline
		\multirow{4}{*}{Combined} & \textit{GloVe-Twitter}+CNN+GRU & 0.40 & 0.49 \\
			& \textit{GloVe-CC}+CNN+GRU & 0.41 & 0.43 \\
			& \textit{FastText-CC}+CNN+GRU & 0.42 & 0.50 \\
			&\textbf{\textit{FastText-Wiki}+CNN+GRU} & 0.42 & 0.52 \\
		\end{tabular}
		\label{tab:weak}
\end{table}

We also experiment with training a set of ensembles using our weak supervision heuristic on posts from Gab itself. For social media where there are no manually labeled datasets available, such as Gab, our experiments demonstrate that weak supervision provides a viable alternative for model training, especially when used in concert with classifier tuning. Table~\ref{tab:weak_gab} illustrates the performance of the Gab classifier ensembles. The best weak supervised ensemble trained on unannotated Gab posts achieved a 62\% hate recall, showing an 8\% improvement after tuning. In all, our weak supervised models are able to achieve consistent performance on hate recall when evaluated on novel data from Gab, surpassing the hate recall of state of the art deep models on the HON dataset. These weak ensembles, as well as the standard pre-trained ensembles trained on the labeled HON and OLID datasets, demonstrate significant improvement in performance from the application of transfer learning by tuning models to an unseen target dataset, such as Gab.

\begin{table*}
	\renewcommand*{\arraystretch}{1.3}
	\centering

	\caption{Comparison of weak supervised ensembles evaluated on the \emph{Gab} dataset. Results are reported before and after tuning is applied to the models. Tuned ensembles showed an average improvement of 12\% on Hate recall }
	\begin{tabular}{cccccc}
		\multicolumn{6}{c}{\textbf{Weak Supervised Ensembles - Gab}} \\
		& & \multicolumn{2}{c}{\textbf{Pre-tuning}} & \multicolumn{2}{c}{\textbf{Post-tuning}} \\
		\textbf{Train Data} & \textbf{Topology} & \textbf{F1-Macro} & \textbf{Hate Recall} & \textbf{F1-Macro} & \textbf{Hate Recall} \\
		\hline
		\multirow{4}{*}{OLID} & \textit{GloVe-Twitter}+CNN+GRU & 0.31 & 0.20 & 0.37 & 0.59 \\
			& \textbf{\textit{GloVe-CC}+CNN+GRU} & \textbf{0.35} & \textbf{0.25} & \textbf{0.40} & \textbf{0.67} \\
			& \textit{FastText-CC}+CNN+GRU & 0.33 & 0.13 & 0.40 & 0.49 \\
			& \textit{FastText-Wiki}+CNN+GRU & 0.35 & 0.23 & 0.39 & 0.54 \\
		\hline
		\multirow{4}{*}{Combined} & \textbf{\textit{GloVe-Twitter}+CNN+GRU} & 0.25 & 0.26 & 0.31 & 0.50 \\
			& \textit{GloVe-CC}+CNN+GRU & 0.42 & 0.27 & 0.32 & 0.42 \\
			& \textit{FastText-CC}+CNN+GRU & 0.27 & 0.26 & 0.34 & 0.39 \\
			& \textit{FastText-Wiki}+CNN+GRU & 0.25 & 0.24 & 0.32 & 0.37 \\
		\hline
		\hline
		\multirow{4}{*}{Gab} & \textit{GloVe-Twitter}+CNN+GRU & 0.31 & 0.51 & 0.36 & 0.53 \\
			& \textit{GloVe-CC}+CNN+GRU & 0.33 & 0.49 & 0.37 & 0.55 \\
			& \textbf{\textit{FastText-CC}+CNN+GRU} & 0.36 & 0.54 & 0.41 & 0.62 \\
			& \textit{FastText-Wiki}+CNN+GRU & 0.32 & 0.61 & 0.38 & 0.60 \\
	\end{tabular}
	\label{tab:weak_gab}
\end{table*}

\section{Conclusions \& Future Work}

In conclusion, in this paper we set out to address the persistent issues of class imbalance and model variability that pose serious challenges for automatic hate speech detection models. Our experiments focus on two major questions regarding how to improve current classification models for automatic hate speech detection and how to develop a generalized framework for hate speech detection that can be extended to unlabeled data. We advance current state of the art models by developing an improved model framework that leverages existing deep topologies to create an ensemble model for automatic hate speech detection Our experiments demonstrate that an ensemble approach outperforms individual models by an average of 5\% hate recall and 8\% F1 macro and is able to improve upon state of the art performance on the HON dataset, achieving 83\% recall on the hate class---a 33\% increase over comparable deep models.

We also show that through the application of transfer learning techniques, we are able to tune our classifier ensembles using a small, curated sample of labeled posts from a particular target dataset. In our experiments, we tune our models trained on the labeled HON and OLID datasets to data from Gab. Tuning improved performance on the Gab dataset by an average of 12\% hate recall, and the best tuned ensemble models achieved 57\% hate recall. We also develop a weak supervision methodology to train and tune our framework entirely on unlabeled data. Our best weak supervised model achieved a 67\% hate recall on Gab.

The vocabulary of hate speech online continues to evolve rapidly along with the nature of social media generally; therefore, adaptable and generalizable automatic techniques for hate speech detection are crucial in order to keep up with the pace of change online. Our simple ensemble and tuning approaches show promise and provide avenues for future work on improving co-training routines and decision making processes for the ensemble, with additional further experimentation to optimize the parameters for both the weak supervision and tuning procedures. Finally, we should seek to develop weak superivsion techniques that are better suited to the evolving nature of hate speech online by forgoing fixed lexicons in favor of more flexible methods.

\section*{Acknowledgments}
The authors thank Michael Ridenhour and the rest of NASCL at UNC-Charlotte for their support in the execution of this work.


\end{document}